\documentclass[sigconf]{acmart}

\copyrightyear{2025}
\acmYear{2025}
\setcopyright{acmlicensed}\acmConference[ICMR '25]{Proceedings of the 2025
International Conference on Multimedia Retrieval}{June 30-July 3,
2025}{Chicago, IL, USA}
\acmBooktitle{Proceedings of the 2025 International Conference on Multimedia
Retrieval (ICMR '25), June 30-July 3, 2025, Chicago, IL, USA}
\acmDOI{10.1145/3731715.3733315}
\acmISBN{979-8-4007-1592-1/25/06}

\begin{document}

\title{DOPE: Dual Object Perception-Enhancement Network for Vision-and-Language Navigation}


\author{Yinfeng Yu}
\orcid{0000-0003-3089-4140}
\affiliation{%
 \department{School of Computer Science and Technology}
  \institution{Xinjiang University}
  \city{Urumqi}
  \country{China}
}
\email{yuyinfeng@xju.edu.cn}

\author{Dongsheng Yang}
\orcid{0009-0000-5799-4307}
\authornote{Corresponding author.}
\affiliation{%
 \department{School of Computer Science and Technology}
  \institution{Xinjiang University}
  \city{Urumqi}
  \country{China}
}
\email{917737683@qq.com}







%
\renewcommand{\shortauthors}{Yinfeng Yu and Dongsheng Yang}

%
\begin{abstract}
Vision-and-Language Navigation (VLN) is a challenging task where an agent must understand language instructions and navigate unfamiliar environments using visual cues. The agent must accurately locate the target based on visual information from the environment and complete tasks through interaction with the surroundings. Despite significant advancements in this field, two major limitations persist: (1) Many existing methods input complete language instructions directly into multi-layer Transformer networks without fully exploiting the detailed information within the instructions, thereby limiting the agent's language understanding capabilities during task execution; (2) Current approaches often overlook the modeling of object relationships across different modalities, failing to effectively utilize latent clues between objects, which affects the accuracy and robustness of navigation decisions. We propose a Dual Object Perception-Enhancement Network (DOPE) to address these issues to improve navigation performance. First, we design a Text Semantic Extraction (TSE) to extract relatively essential phrases from the text and input them into the Text Object Perception-Augmentation (TOPA) to fully leverage details such as objects and actions within the instructions. Second, we introduce an Image Object Perception-Augmentation (IOPA), which performs additional modeling of object information across different modalities, enabling the model to more effectively utilize latent clues between objects in images and text, enhancing decision-making accuracy. Extensive experiments on the R2R and REVERIE datasets validate the efficacy of the proposed approach.
\end{abstract}

\begin{teaserfigure}
    \includegraphics[width=\textwidth]{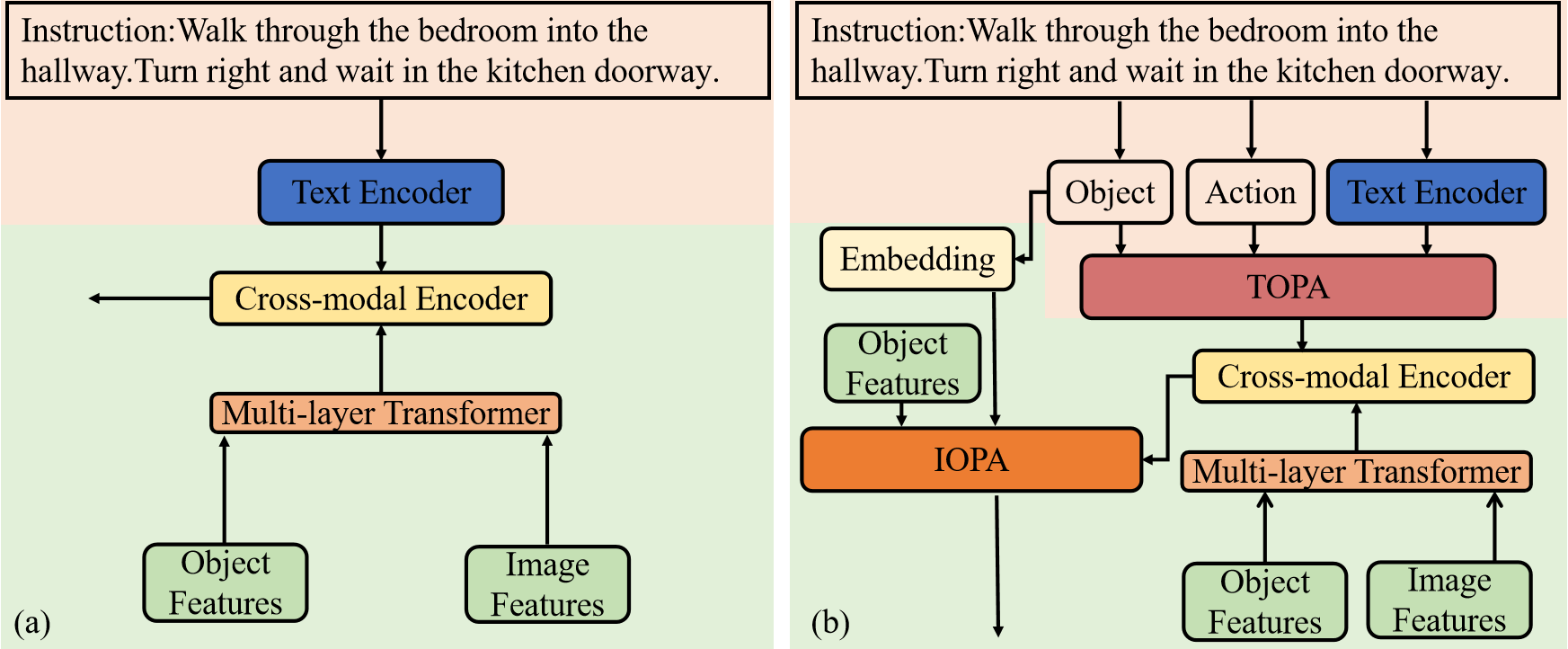} 
    \caption{Illustration of the differences between DOPE and other methods. (a) Conventional methods, (b) Our method.}
    \label{fig:duibitu}
\end{teaserfigure}


\maketitle
\section{Introduction}
With the development of multimodal technologies, embodied intelligence tasks \cite{SAAVN, gupta2017cognitive, datta2022episodic} have gained widespread attention. In particular, navigation tasks have gradually expanded from visual navigation \cite{gupta2017cognitive} to audio-visual navigation \cite{E3VN,YinfengIJCAI2023MACMA,fsaavn} and Vision-and-Language Navigation (VLN) tasks \cite{anderson2018vision, qi2020reverie, qiao2022hop, an2024etpnav}, with VLN emerging as an essential challenge in the field of human-machine interaction. 
The core issue of VLN tasks lies in efficiently integrating language instructions with visual information from the environment to make accurate navigation decisions. The complexity is in understanding the high-level semantic information in the language instructions and fully utilizing the implicit clues in the visual information to achieve effective goal localization and task completion.

In these tasks, \cite{anderson2018vision, zhu2020vision} require the agent to follow step-by-step instructions to reach a specific destination, while \cite{qi2020reverie, liu2023bird} demand that the agent not only reach the destination based on descriptions of the target location but also identify the target object. Therefore, the agent must fully leverage both instructional and environmental information to accomplish these complex tasks.

In recent years, with the successful application of the Transformer architecture \cite{vaswani2017attention} across various domains, significant progress has been made in VLN tasks \cite{  gao2023adaptive,zhou2024navgpt}. However, despite these methods performing excellently on multiple public datasets, numerous challenges remain in practical applications, primarily in the following two aspects.

Firstly, as shown in the pink section of Figure \ref{fig:duibitu} (a), existing methods \cite{li2023kerm, wang2023gridmm, zhou2024navgpt} directly input complete natural language instructions into multi-layer Transformer networks without fully mining the detailed information within the instructions. This approach somewhat limits the agent’s deep understanding of language instructions, especially during the execution of complex tasks, where it is prone to overlook key information in the instructions, such as specific actions and target objects, thereby affecting the accuracy of navigation decisions.

Secondly, as shown in the green section of Figure \ref{fig:duibitu} (a), existing methods \cite{chen2022think} model cross-modal relationships by directly combining instruction features with image features that include object characteristics from visual information. These methods often neglect modeling the relationships between objects across different modalities. There are inherent associations between objects in visual information and those in language instructions, which contain rich cues that help the agent more accurately understand the environment and task requirements. However, existing models often fail to effectively utilize these inter-object relationships during multimodal information fusion, thus impacting overall navigation performance.

To address the abovementioned issues, we propose the Dual Object Perception-Enhancement Network (DOPE), aimed at enhancing navigation performance in VLN tasks by deepening the understanding of language instruction details and modeling multimodal object relationships. Specifically, DOPE achieves this goal through the collaborative operation of three key modules. Firstly, the Text Semantic Extraction (TSE) extracts critical information, such as action verbs and target object nouns, from natural language instructions to ensure that the agent accurately comprehends the core content of the instructions. Next, the Text Object Perception-Augmentation (TOPA) further processes the extracted key information using a multi-head attention mechanism to enhance object and action information within the instructions, thereby deepening language understanding. Simultaneously, the Image Object Perception-Augmentation (IOPA) conducts in-depth mining of objects within visual information, using a cross-modal encoder to model object relationships across different modalities, thereby enhancing environmental understanding and navigation decision accuracy. These three modules' synergistic effect enables DOPE to combine language and visual information more effectively, improve object perception capabilities, and significantly enhance navigation performance.

As shown in the pink section of Figure \ref{fig:duibitu} (b), when processing textual information, object and action information is first extracted and then combined with the original instructions as input to the TOPA module to enhance object perception. In addition, as shown in the green section of Figure 1 (b), when processing visual information, the proposed IOPA module models object features from both image and text, and these modeled results are combined with the fully modeled results and input into the IOPA to obtain enhanced cross-modal information.

Through extensive experiments on the publicly available R2R \cite{anderson2018vision} and REVERIE \cite{qi2020reverie} datasets, this paper validates the effectiveness of the DOPE method in enhancing navigation performance. Experimental results indicate that DOPE outperforms existing methods in metrics such as task completion rate and path efficiency and demonstrates higher robustness when faced with complex environments and instructions.This paper makes the following key contributions:
\begin{enumerate}
    \item We propose a Dual Object Perception-Enhancement Network (DOPE), which effectively enhances the depth of language instruction understanding and the modeling of multimodal object relationships in VLN tasks.
    
    \item We design the Text Semantic Extraction (TSE) and Text Object Perception-Augmentation (TOPA), which enhance the utilization of critical information in instructions through fine-grained language processing.
    
    \item We introduce the Image Object Perception-Augmentation (IOPA), which utilizes a cross-modal encoder to deeply mine object relationships between text and visual information, improving the accuracy and robustness of navigation decisions.
    
    \item We conduct comprehensive experimental evaluations on the R2R and REVERIE datasets, demonstrating that the DOPE method outperforms existing methods across multiple metrics.
\end{enumerate}

\section{Related Work}
\begin{figure*}[t]
    \centering
    \includegraphics[width=\textwidth]{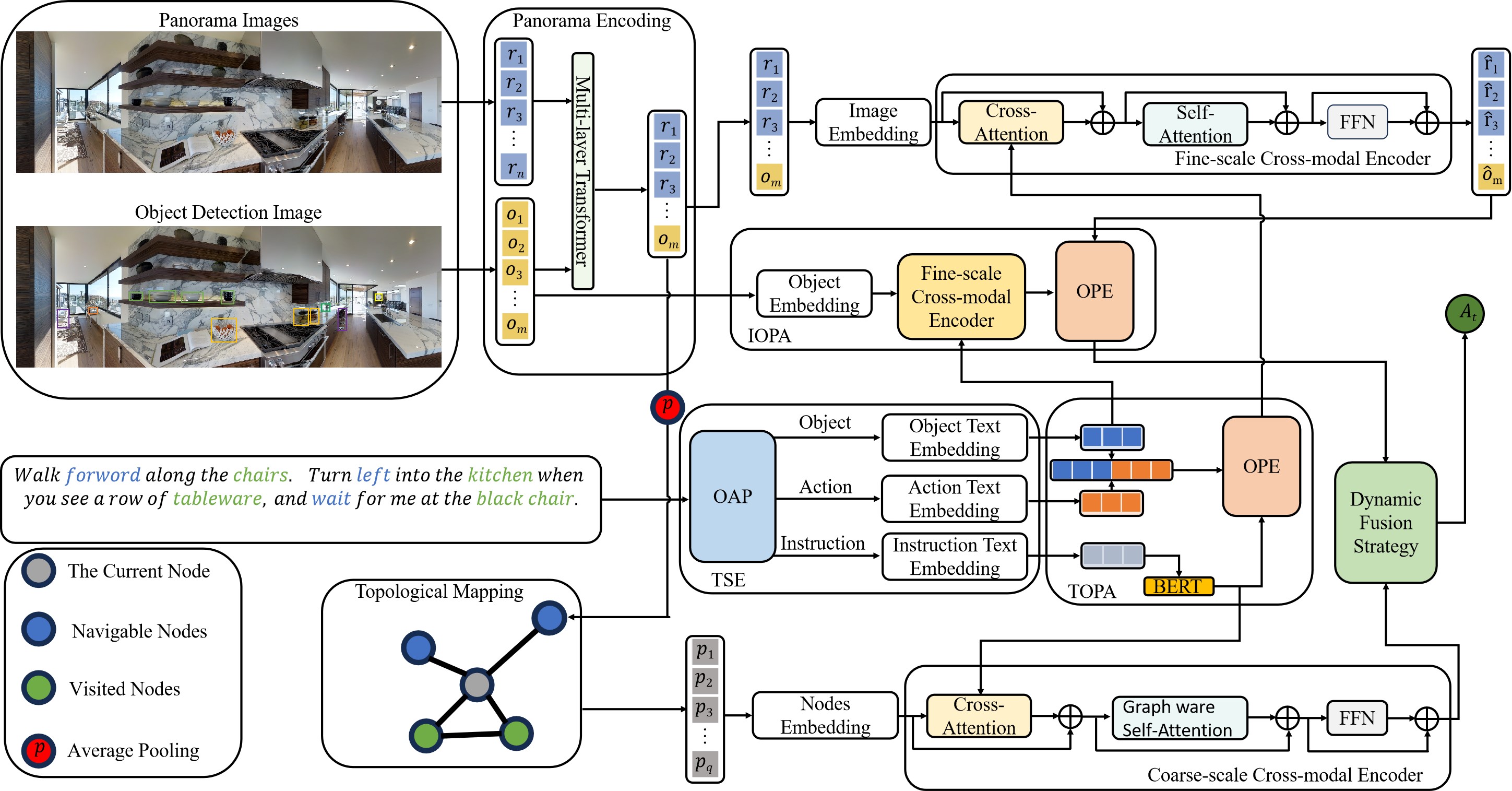} 
    \caption{Overview of the proposed Dual Object Perception-Enhancement Network (DOPE), which primarily comprises three key components: Text Semantic Extraction (TSE), Text Object Perception Augmentation (TOPA), and Image Object Perception Augmentation (IOPA).}
    \label{fig:kjt}
\end{figure*}
\subsection{Vision-and-Language Navigation}
Vision-and-Language Navigation (VLN) tasks \cite{anderson2018vision} have garnered substantial research interest since their inception \cite{lin2023learning,he2024frequency} and have achieved remarkable progress with the emergence of new technologies. Early models primarily utilized Recurrent Neural Networks (RNNs) to encode relevant information \cite{anderson2018vision,fried2018speaker,tan2019learning}. In recent years, benefiting from the successful application of Transformer architectures \cite{vaswani2017attention} across various domains, Transformer-based architectures have made significant strides in VLN tasks \cite{chen2022think,schumann2024velma}.

To better leverage historical information, HAMT \cite{chen2021history} captured long-term dependencies of all past observations and actions by directly employing Transformers.\cite{wen2024vision} proposed a history-aware VLN method based on cross-modal feature fusion. To enhance model generalization capabilities, environment augmentation methods were proposed \cite{li2022envedit,wang2023res} to increase environmental diversity.Additionally, introducing auxiliary tasks \cite{devlin2018bert,lu2019vilbert,chen2021history,lin2021scene} were shown to improve model performance.Although these works have made good progress, there are still shortcomings in utilizing object information from instructions and images. Therefore, we propose a Dual Object Perception-Enhancement model to enable the agent to leverage the available information better.
\subsection{Semantic Information in VLN}
Semantic features are pivotal in VLN tasks, facilitating the agent's ability to accurately recognize and understand the detailed information in the environment, which in turn enables more efficient navigation. Recently, a number of strategies have been introduced to fully exploit semantic features to improve the performance of VLN tasks.

ORIST \cite{qi2021road} enhanced the model's comprehensive understanding of visual information by concatenating object-level and scene-level image features and using parallel modeling in the encoder.SEvol \cite{chen2022reinforced} further enhances navigation performance by constructing a graph framework that captures the relationships among objects. Although these methods have improved model performance, they primarily focus on the visual aspect and have not fully integrated the key information from language instructions.
On the other hand, OAAM \cite{qi2020object} introduced two attention modules that focused on the object and action information in the instructions, highlighting the relevant semantic details and enhancing the depth and accuracy of language understanding.Unlike the aforementioned methods that only process a single modality, the method introduced in this paper approaches from both visual and linguistic modalities. In terms of language, we first extract object and action information from the text and then combine the extracted data with the original instruction, significantly enhancing the model's ability to understand the key details in the instruction. In terms of vision, we separately extract object features and additionally model the target information in the text to improve the model's understanding of details.
\subsection{Object Relationships in VLN}
Some recent studies have focused on leveraging object relationships to enhance model performance. ORG \cite{du2020learning} improved the learning of visual representations by integrating relationships between objects, such as category proximity and spatial relevance. CKR \cite{gao2021room} utilized common-sense knowledge from external knowledge bases to learn the associations between rooms and objects, enhancing the model's understanding of the environment. KERM \cite{li2023kerm} built an external knowledge base to assist in establishing relationships between various entities in the instruction, thereby improving decision-making in navigation tasks.

Compared to previous methods, the approach proposed in this paper models the object relationships in both modalities (visual and language) separately, learning the relationships between objects in depth and enhancing the model's perception of objects. This method avoids biases that external knowledge bases may introduce and enables the model to more efficiently extract and leverage key information from both the environment and the instruction. By comprehensively modeling object relationships across both modalities, the proposed method notably enhances the accuracy of navigation decisions.
\begin{figure}[t]
    \centering
    \includegraphics[width=\linewidth]{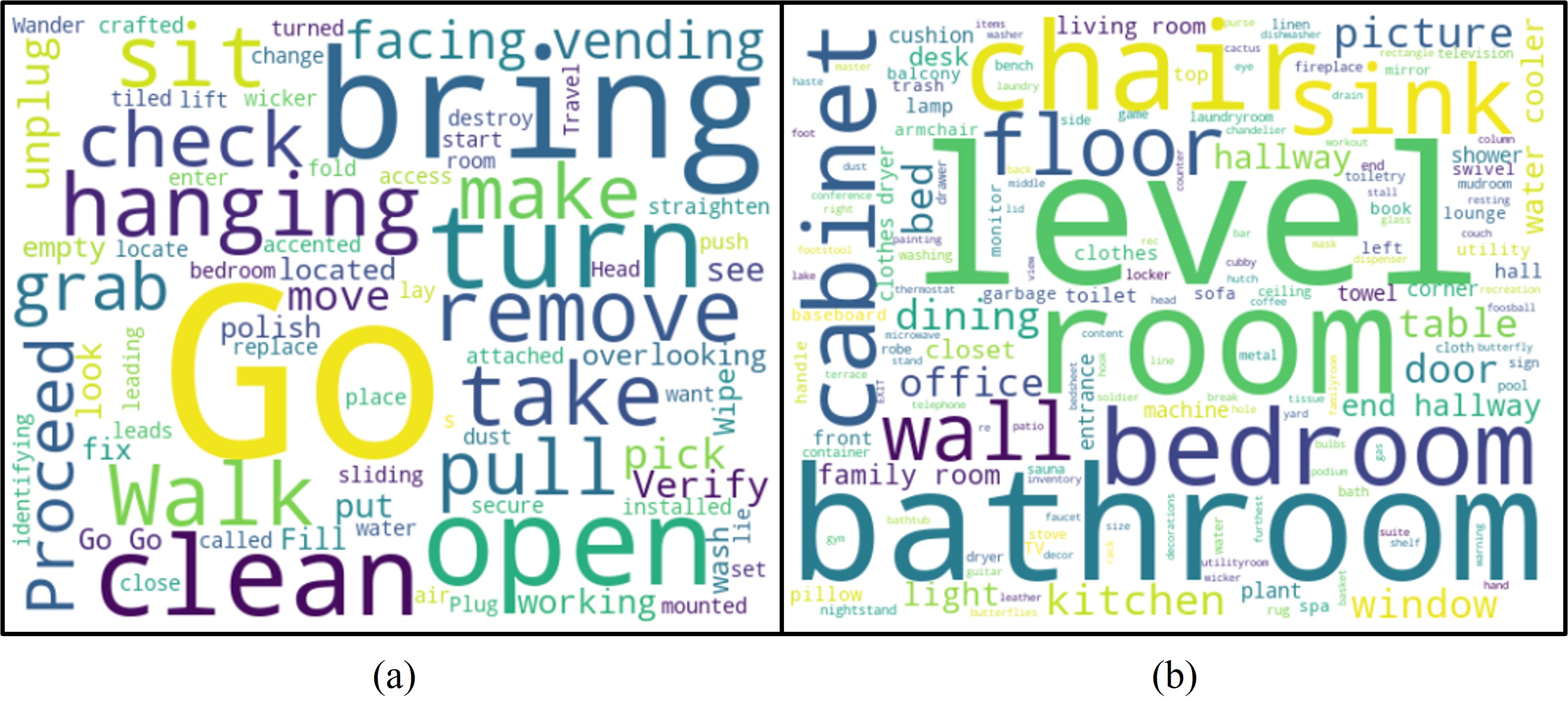}
    \caption{(a) Action Word Cloud and (b) Object Word Cloud.}
    \label{fig:word_clouds}
\end{figure}
\section{Method}
We conduct the task in a discrete indoor environment, which is represented by an undirected graph $G = \{V, E\}$, where $V = \{V_i\}_{i=1}^K$ denotes $K$ navigable nodes, and $E$ represents the edges connecting these nodes. During the initialization phase, the agent is equipped with a GPS sensor and an RGB camera, and is positioned at a starting node in an unseen environment. The instructions are embedded as $I = \{w_i\}_{i=1}^L$, where $L$ denotes the number of words in the instructions. At each time step $t$, the agent observes a panoramic view and receives related positional information of the current node $V_t$. The panoramic view is divided into $n$ viewpoint images, denoted as $R_t = \{r_i\}_{i=1}^n$, where $r_i$ represents the image features of the $i$-th viewpoint. While navigating, the agent must choose the subsequent action $a_t$ from the action set $A_t = \{a_{t, i}\}_{i=0}^k$ to navigate to the next node. This decision-making process persists until the agent arrives at the destination specified in the instructions or exceeds the maximum number of allowed steps. For target-oriented tasks \cite{qi2020reverie,zhu2021soon}, the agent must also localize and find the target object at the target location.

\subsection{Method Overview}

The objective of the VLN task is to predict a sequence of actions based on natural language instructions and environmental information to complete the navigation. The core objective of this study is to fully utilize the object information in the visual representations and instructions to enhance the model's navigation ability significantly. To achieve this, we choose the DUET \cite{chen2022think} as the baseline, which has performed excellently in VLN tasks.

The DUET model receives three main inputs: the panoramic view at time step $t$, the dynamically updated topological map, and the natural language instructions. The nodes in the topological map are categorized into three types: current nodes, visited nodes, and navigable nodes, representing the agent's position and reachable paths in the environment. As shown in Figure \ref{fig:kjt}, based on the DUET\cite{chen2022think}, we propose an object perception-enhancement approach, which consists of three modules: (a) Text Semantic Extraction (TSE), (b) Text Object Perception-Augmentation (TOPA), and (c) Image Object Perception-Augmentation (IOPA). These three modules collaborate to form a comprehensive semantic perception system, improving the model's understanding of instructions and visual information and enhancing navigation performance.
\subsection{Text Semantic Extraction}
\label{subsec:tse}
In Vision-and-Language Navigation (VLN) tasks, natural language instructions typically contain much information, but not all words are equally crucial for task execution. To enhance the model's understanding of language instructions, particularly the accuracy of navigation decisions, we focus on the key informational components within the instructions, such as nouns and verbs. Nouns generally represent the target objects in the task, while verbs indicate actions or behaviors. By effectively identifying and processing these critical words, the agent can more precisely understand the targets and operations described in the instructions, thereby optimizing its navigation strategy.

\subsubsection{Object and Action Parser}
\label{subsubsec:oap}

As shown in Figure \ref{fig:word_clouds} (a) and (b), the Object and Action Parser (OAP) module is designed to extract two types of key semantic information from natural language instructions: action verbs and target object nouns. To achieve this, we employ a pre-trained DistilBERT tokenizer \cite{sanh2019distilbert} and spaCy's language model for tokenization and part-of-speech tagging of the input instructions. Specifically, the Object and Action Parser (OAP) module first defines an action vocabulary relevant to the navigation task to identify action verbs within the instructions. Additionally, for target object nouns tagged as nouns, the module performs normalization through regular expression cleaning, lemmatization, and removal of numbers to ensure consistency and accuracy in subsequent processing.

After processing with the Object and Action Parser (OAP), we obtain object phrases \( \mathcal{I}_o = \{w_1^o, w_2^o, \ldots, w_{o_L}^o\} \), action phrases \( \mathcal{I}_a = \{w_1^a, w_2^a, \ldots, w_{a_L}^a\} \), and the original instruction \( \mathcal{I} = \{w_1, w_2, \ldots, w_{L}\} \). Subsequently, each word in these three data sets is embedded into a 768-dimensional vector space. To preserve the sequential information of the words, we inject the sequence information into the word vectors through a positional embedding function, as illustrated in Equation (\ref{eq1}).
\begin{equation}
    (\mathcal{I}, \mathcal{I}_o, \mathcal{I}_a)= \text{Embedding}(\mathcal{I}, \mathcal{I}_o, \mathcal{I}_a). \label{eq1}
\end{equation}
\subsection{Text Object Perception-Augmentation}
\label{subsec:topa}

After processing by the TSE module, the instructions are input into the Text Object Perception-Augmentation (TOPA) for further processing to obtain object perception-enhanced instructions. Specifically, the TOPA module first concatenates the action embeddings processed with positional embeddings with the object embeddings. Subsequently, it employs the pre-trained BERT model \cite{devlin2018bert} to encode the words in the instructions, thereby obtaining contextual language features, denoted as \( \mathcal{E}_c^I \).

\begin{figure}[t]
    \centering
    \includegraphics[width=\linewidth]{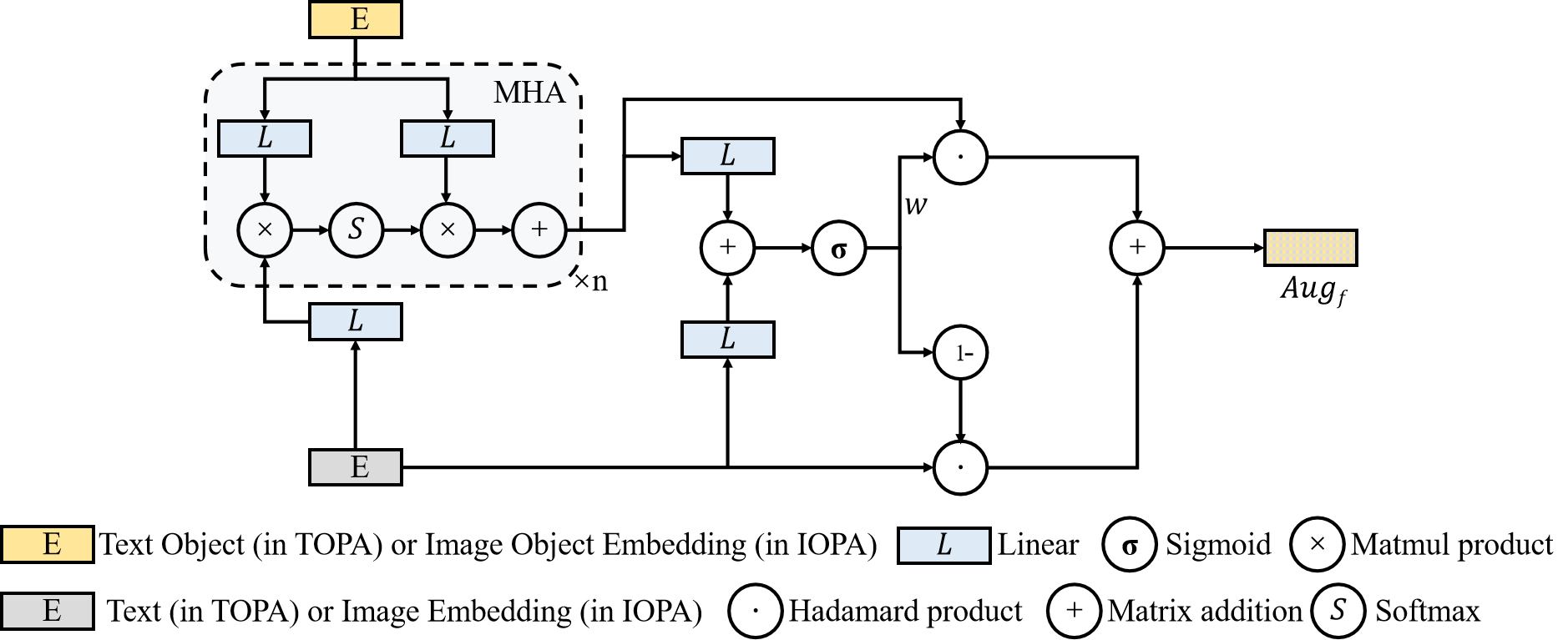}  
    \caption{Illustration of the Object Perception-Enhancement module.}
    \label{fig:OPE}
\end{figure}

To emphasize the importance of object and action phrases in the text, we design an Object Perception-Enhancement (OPE) module within the TOPA module, as illustrated in Figure \ref{fig:OPE}. The OPE module updates the relationships between the contextual features and the object phrases and action phrases by introducing a Multi-Head Attention (MHA) mechanism \cite{vaswani2017attention}, thereby obtaining enhanced contextual language features, denoted as \( \mathcal{E}_g^I \). Subsequently, to dynamically balance the ratio between \( \mathcal{E}_c^I \) and \( \mathcal{E}_g^I \), we adopt a gate-like structure, dynamically computing weight values $w$ using a sigmoid function. These weights are then used to perform a weighted sum of \( \mathcal{E}_c^I \) and \( \mathcal{E}_g^I \), resulting in the enhanced feature \( \mathcal{A}_f^I \). The following equations represent the above process:
\begin{align}
    \mathcal{E}_c^{I} &= \text{BERT}(\mathcal{I}), \label{eq2} \\
    \mathcal{E}_g^{I} &= \text{MHA}(\mathcal{E}_c^{I}, [\mathcal{I}_o, \mathcal{I}_a]), \label{eq3} \\
    \omega &= \delta (\mathcal{E}_g^{I} W_g + \mathcal{E}_c^{I} W_c + b_I), \label{eq4} \\
    \mathcal{A}_f^I &= \omega \odot \mathcal{E}_g^{I} + (1 - \omega) \odot \mathcal{E}_c^{I}. \label{eq5}
\end{align}
\subsection{Image Object Perception-Augmentation}
\label{subsec:iopa}

\begin{figure}[t]
    \centering
    \includegraphics[width=\linewidth]{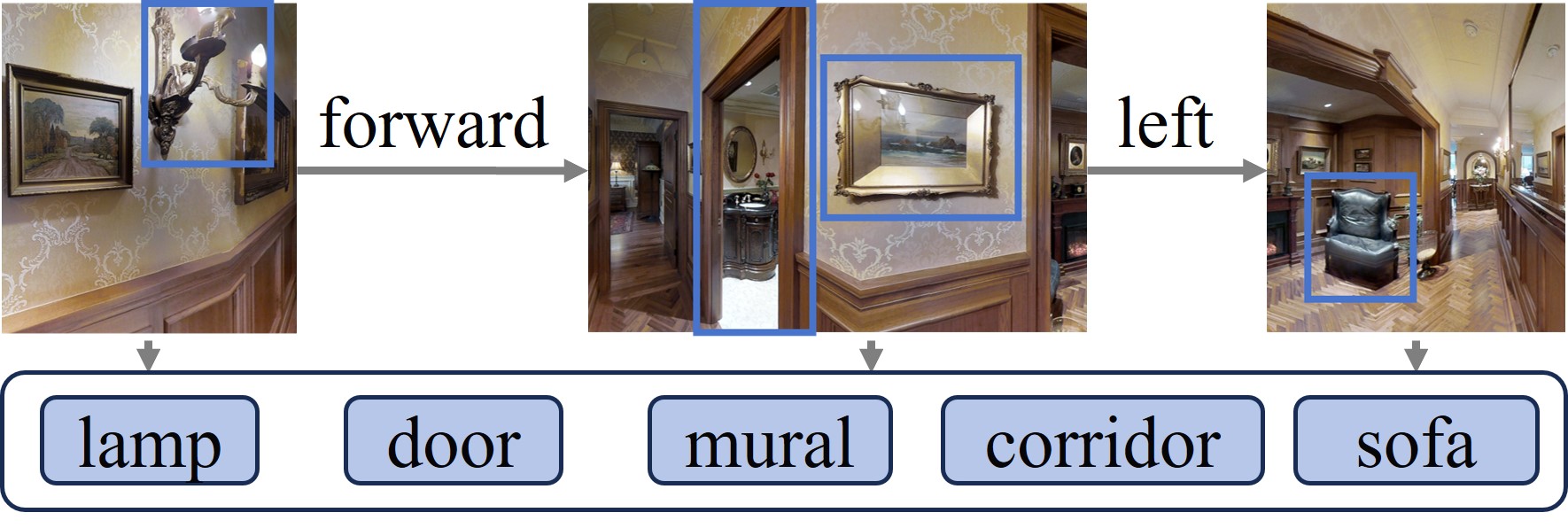}
    \caption{Illustration of cross-modal object modeling.}
    \label{fig:duiqi}
\end{figure}
\begin{table*}[t]
\centering
\caption{Comparison with the state-of-the-art methods on the R2R dataset}
\label{tab:r2r}
\resizebox{0.97\textwidth}{!}{%
\begin{tabular}{lccccccccccccc}
\toprule
Methods & \multicolumn{4}{c}{Val Seen} & \multicolumn{4}{c}{Val Unseen} & \multicolumn{4}{c}{Test Unseen} \\
\cmidrule(lr){2-5} \cmidrule(lr){6-9} \cmidrule(lr){10-13}
 & NE$\downarrow$ & OSR$\uparrow$ & SR$\uparrow$ & SPL$\uparrow$
 & NE$\downarrow$ & OSR$\uparrow$ & SR$\uparrow$ & SPL$\uparrow$
 & NE$\downarrow$ & OSR$\uparrow$ & SR$\uparrow$ & SPL$\uparrow$ \\ 
\midrule
EnvDrop \cite{tan2019learning} & 3.99 & - & 62 & 59 & 5.22 & - & 52 & 48 & 5.23 & 59 & 51 & 47 \\ 
EnvEdit \cite{li2022envedit} & 2.17 & - & 77 & 74 & 3.24 & - & 69 & \textbf{64} & 3.59 & - & 68 & 64 \\  
HOP+ \cite{qiao2023hop+} & 2.33 & - & 78 & 73 & 3.49 & - & 67 & 61 & 3.71 & - & 66 & 60 \\ 
DUET \cite{chen2022think} & 2.28 & 86 & 79 & 73 & 3.31 & 81 & 72 & 60 & 3.65 & 76 & 69 & 59 \\ 
KERM \cite{li2023kerm} & 2.19 & - & 80 & 74 & 3.22 & - & 72 & 61 & 3.61 & - & 70 & 59 \\ 
GridMM \cite{wang2023gridmm} & 2.34 & - & 80 & 74 & 2.83 & - & \textbf{75} & \textbf{64} & 3.35 & - & 73 & 62 \\ 
LSAL \cite{wu2024vision} & 2.88 & - & 73 & 70 & 3.62 & - & 65 & 59 & 4.00 & - & 63 & 58 \\ 
NavGPT2 \cite{zhou2024navgpt} & 2.84 & 83 & 74 & 63 & 2.98 & \textbf{84}  & 74 & 61 & 3.33 & 80 & 72 & 60 \\ 
KESU \cite{gao2024enhancing} & 2.19 & - & \textbf{81} & \textbf{75} & 2.96 & - & 73 & 62 & 3.31 & - & 72 & 61 \\
BEVBert \cite{an2023bevbert} & 2.17 & - & \textbf{81} & 74 & 2.81 & - & \textbf{75} & \textbf{64} & 3.13 & \textbf{81} & 73 & 62 \\ 
\textbf{DOPE(Ours)} & \textbf{2.10} & \textbf{89} & \textbf{81} & \textbf{75} & \textbf{2.80} & \textbf{84} &  \textbf{75} & \textbf{64} & \textbf{3.06} & \textbf{81} & \textbf{74} & \textbf{63} \\
\bottomrule
\end{tabular}
}
\end{table*}

In Vision-and-Language Navigation (VLN) tasks, another key input modality is the environmental information observed visually, particularly the objects within the environment, which play a crucial role during navigation. To fully utilize this object information, we propose an Image Object Perception-Augmentation (IOPA) to boost the model's object perception capabilities in the environment.Through this method, the model can effectively identify important objects closely related to the task instructions, such as landmarks at target locations or key points along the navigation path, thereby improving the overall understanding of the environment.

At time step $t$, for the input panoramic images and image objects, we use CLIP \cite{radford2021learning} to extract their features, denoted as $R_t$ and $O_t$, respectively. Subsequently, we use a multi-layer Transformer architecture to model the spatial relationships between images and objects. The following equations represent the above steps:
\begin{align}
    R_t, O_t &= \operatorname{CLIP}(\mathcal{R}_t, \mathcal{O}_t), \label{eq6} \\
    [R'_t, O'_t] &= \operatorname{SelfAttn}([R_t, O_t]) \label{eq7} , \\
    \operatorname{SelfAttn}(X) &= \operatorname{Softmax}\left(\frac{X W_q (X W_k)^\top}{\sqrt{d}}\right) X W_v . \label{eq8}
\end{align}
\subsubsection{Object Embedding}
\label{subsubsec:object_embedding}

First, at time step $t$, we process the object features $O_t$ and add two types of positional embeddings. The first type of embedding represents the current node's position in relation to the starting node, which not only helps the model understand the spatial localization in the instructions but also enhances the model's comprehension of spatial relationships in the environment. To further assist the agent in understanding egocentric directional information, we introduce a second type of positional embedding that indicates the relative positions of neighboring nodes to the current node.

\subsubsection{Fine-grained Cross-modal Encoder}
\label{subsubsec:cross_modal_encoder}

After obtaining the object features $O_t$ and language object embeddings $\mathcal{I}_o$, we learn the correlations between these two types of features by integrating them, as shown in Figure \ref{fig:duiqi}. Specifically, we employ the LXMERT model \cite{tan2019lxmert} as the cross-modal encoder to model the relationships between object features and language object features, thereby obtaining the integrated object features $\overline{O}_t = \{\overline{o}_i\}_{i=1}^m$.

For the panoramic features, we similarly add the two types of positional embeddings to the obtained $[R'_t, O'_t]$ features and finally input them together with the object-enhanced text features processed by TOPA (Text Object Perception-Augmentation) into the fine-grained cross-modal encoder to model the relationships between visual features and the enhanced text features. The output of this process is represented as $f_t = \{\hat{R}_t, \hat{O}_t\}$.

We input $\overline{O}_t$ and $f_t$ into the proposed IOPA module to enhance the model's image object perception capability. By using a Multi-Head Attention (MHA) mechanism, we update the relationships between $\overline{O}_t$ and $f_t$, thereby obtaining the object-enhanced image features. Subsequently, to balance the proportion between the object-enhanced image features and $\mathcal{E}_g^I$, we adopt a gate-like structure to obtain the final image features $\mathcal{A}_f^R$.

\subsection{Dynamic Fusion Strategy}
\label{subsec:dynamic_fusion}

We predict the model's next action following the DUET \cite{chen2022think} approach during the action selection process. Specifically, for each topological node feature $\overline{p}_t = \{\overline{p}_i\}_{i=1}^q$ processed by the Coarse-scale Cross-modal Encoder, we predict a global action space navigation score $s_i^c$, represented as:
\begin{equation}
    S_i^c= \mathrm{FFN}(\hat{p}_i). \label{eq9}
\end{equation}
FFN denotes a two-layer feedforward network.

For the obtained $\mathcal{A}_f^R$, we similarly predict a local action score $S_i^f$ according to Equation (\ref{eq9}), and then convert $S_i^f$ into a global action space navigation score $\hat{S}_i^c$. Subsequently, we perform a weighted fusion of $S_i^c$ and $\hat{S}_i^c$ to obtain the final action prediction probability $S_i$ and select the action with the highest score for the next navigation step.

\section{Experiments}
\begin{table*}[t]
\centering
\caption{Comparison with the state-of-the-art methods on the REVERIE dataset}
\label{tab:reverie}
\resizebox{0.97\textwidth}{!}{%
\begin{tabular}{lcccccccccc}
\toprule
Methods& \multicolumn{5}{c}{Val Unseen} & \multicolumn{5}{c}{Test Unseen} \\
\cmidrule(lr){2-6} \cmidrule(lr){7-11}
& OSR$\uparrow$ & SR$\uparrow$ & SPL$\uparrow$ & RGS$\uparrow$ & RGSPL$\uparrow$ 
& OSR$\uparrow$ & SR$\uparrow$ & SPL$\uparrow$ & RGS$\uparrow$ & RGSPL$\uparrow$ \\
\midrule
Airbert\cite{guhur2021airbert} & 34.51 & 27.89 & 21.88 & 18.23 & 14.18 & 34.20 & 30.28 & 23.61 & 16.83 & 13.28 \\ 
HOP+\cite{qiao2023hop+} & 40.04 & 36.07 & 31.13 & 22.49 & 19.33 & 35.81 & 33.82 & 28.24 & 20.20 & 16.86 \\ 
DUET\cite{chen2022think} & 51.07 & 46.98 & 33.73 & 32.15 & 23.03 & 56.91 & 52.51 & 36.06 & 31.88 & 22.06 \\ 
LANA \cite{wang2023lana} & 52.97 & 48.31 & 33.86 & 32.86 & 22.77 & 57.20 & 51.72 & 36.45 & 32.95 & 22.85 \\ 
KERM\cite{li2023kerm} & 55.21 & 49.02 & 34.83 & 33.97 & 24.14 & 57.44 & 52.26 & 37.46 & 32.69 & 23.15 \\ 
FDA\cite{he2024frequency} & 51.41 & 47.57 & 35.90 & 32.06 & 24.31 & 53.54 & 49.62 & 36.45 & 30.34 & 22.08 \\ 
CONSOLE\cite{lin2024correctable} & 54.25 & 50.07 & 34.40 & 34.05 & 23.33 & 59.60 & 55.13 & 37.13 & 33.18 & 22.25 \\ 
ACME\cite{wu2025adaptive} & 53.97 & 49.46 & 32.37 & 32.64 & 24.02 & 57.48 & 51.89 & 34.65 & 33.12 & 23.57 \\ 
BEVBert\cite{an2023bevbert} & 56.40 & \textbf{51.78} & \textbf{36.37} & 34.71 & 24.44 & 57.26 & 52.81 & 36.41 & 32.06 & 22.09 \\ 
ACK\cite{mohammadi2024augmented} & 52.77 & 47.49 & 34.44 & 32.66 & 23.92 & 59.01 & 53.97 & 37.89 & 32.77 & 23.15 \\ 
\textbf{DOPE(Ours)} & \textbf{56.92} & 51.72 & 36.11 & \textbf{35.27} & \textbf{24.97} & \textbf{63.10} & \textbf{58.38} & \textbf{41.87}  & \textbf{35.84} & \textbf{25.43} \\
\bottomrule
\end{tabular}
}
\end{table*}
\begin{table}[t]
\centering
\caption{Ablation Study of Object Enhancement Modules}
\label{tab:all}
\begin{tabular}{ccccccc} 
\toprule
Id & TSE+TOPA & IOPA & SR$\uparrow$ & SPL$\uparrow$ & RGS$\uparrow$ & RGSPL$\uparrow$ \\
\midrule
1  & $\times$ & $\times$ & 46.98 & 33.73 & 32.15 & 23.03 \\ 
2  & $\checkmark$ & $\times$ & 49.47 & 33.74 & 33.57 & 23.38 \\ 
3  & $\times$ & $\checkmark$ & 50.01 & 34.17 & 34.73 & 23.95 \\ 
4  & $\checkmark$ & $\checkmark$ & \textbf{51.72} & \textbf{36.11} & \textbf{35.27} & \textbf{24.97} \\ 
\bottomrule
\end{tabular}
\end{table}

\begin{table}[t]
\caption{Ablation Study of Object Perception-Enhancement Modules}
\label{tab:past}
\resizebox{\columnwidth}{!}{  
\begin{tabular}{lcccccc}
\toprule
Method & Configuration & SR$\uparrow$ & SPL$\uparrow$ & RGS$\uparrow$ & RGSPL$\uparrow$ \\
\midrule
TSE+TOPA
    & w/o OPE    & 49.28 & 31.59 & \textbf{34.19} & 21.97 \\ 
    & w/ OPE & \textbf{49.47}  & \textbf{33.74} & 33.57 & \textbf{23.38} \\ 
\midrule
IOPA
    & w/o OPE    & 49.73 & 31.81 & 33.80  & 22.03 \\ 
    & w/ OPE & \textbf{50.01} & \textbf{34.17} & \textbf{34.73} & \textbf{23.95} \\ 
\midrule
ALL
    & w/o OPE    & 47.37 & 31.36 & 33.12 & 22.16 \\ 
    & w/ OPE & \textbf{51.72} & \textbf{36.11} & \textbf{35.27} & \textbf{24.97} \\ 
\bottomrule
\end{tabular}
}
\end{table}
\subsection{Datasets}

To validate the effectiveness of our proposed method, we conducted extensive experiments on two datasets: R2R \cite{anderson2018vision}and REVERIE\cite{qi2020reverie} .
The R2R dataset comprises 90 scenes and 21,567 navigation instructions. The instruction type is path-oriented.The path
lengths range from 4 to 7 steps
The REVERIE dataset consists of 21,702 high-level navigation instructions that describe target locations, with each instruction averaging 21 words and a mean path length of 6 steps. Each panoramic image provides predefined object bounding boxes, and at the end of navigation, the agent is required to select the correct object bounding box. On average, each panoramic sub-image contains about 10 objects.
\subsection{Evaluation Metrics}
We adopt commonly used evaluation metrics in VLN tasks to assess our model's performance and compare it with existing works. On the R2R dataset, we utilize four standard metrics: the Navigation Error (NE), which measures the average distance (in meters) between the agent’s stopping position and the actual target position; the Success Rate (SR), which calculates the proportion of paths where the agent’s stopping position is within 3 meters of the actual target position; the Oracle Success Rate (OSR), which represents the SR under the Oracle stopping strategy; and the Success Rate weighted by Path Length (SPL), which combines success rate and path efficiency by penalizing path length. On the REVERIE dataset, in addition to the metrics above, we also employ two additional metrics: the Remote Grounding Success Rate (RGS), which denotes the proportion of cases where the agent correctly localizes the target object at the target location, and the Remote Grounding Success Rate weighted by Path Length (RGSPL), which combines RGS with path length consideration.
\subsection{Implementation Details}
\label{subsec:implementation_details}
Our model consists of 9 layers of a text Transformer, two layers of a panorama Transformer, and coarse- and fine-grained cross-modal encoders, with the latter two using four layers of Transformer each. Other hyperparameter settings are the same as those in LXMERT \cite{tan2019lxmert}. For image feature extraction, we use CLIP-B/16 \cite{radford2021learning}. Object bounding boxes are provided in the REVERIE dataset, and we also use CLIP for feature extraction. Directional features include the sine and cosine values of the pitch and yaw angles.

During the pretraining phase, we train on the R2R dataset using Masked Language Modeling (MLM) \cite{devlin2018bert}, Masked Region Classification (MRC) \cite{lu2019vilbert}, and Single-step Action Prediction (SAP) \cite{chen2021history}; for the REVERIE dataset, we additionally use Object Grounding (OG) \cite{lin2021scene} for training. To enhance feature representations, we utilize EnvEdit \cite{li2022envedit}. During pretraining, the R2R and REVERIE datasets are trained using 5 Tesla V100 GPUs, with the AdamW optimizer\cite{loshchilov2017decoupled}, and a batch size of 20 per GPU. The maximum number of training iterations for the REVERIE dataset is 100K, while for the R2R dataset, it is 355K. In the fine-tuning phase, a single Tesla V100 GPU is used for the REVERIE and R2R datasets, with batch sizes of 6 and 8, respectively.
\subsection{Comparison with Existing Methods}
\label{subsec:comparison_methods}
Our method is evaluated against prior works on the R2R and REVERIE datasets, with detailed results shown in Table \ref{tab:r2r} and Table \ref{tab:reverie}. On both datasets, our method demonstrates superior navigation accuracy, instruction following precision, and object grounding performance in seen and unseen environments.
In the R2R dataset, our method achieves the best performance across all metrics in both seen and unseen environments compared to previous approaches.
On test unseen splits, compared to the baseline DUET\cite{chen2022think}, our method shows a decrease of 0.59 in the NE metric, while OSR, SR, and SPL are improved by 5\%, 5\%, and 4\%, respectively.
Our method demonstrates significant improvements across all metrics relative to existing methods on the more challenging test set of the REVERIE dataset. Specifically, compared to ACK\cite{mohammadi2024augmented}, our approach enhances OSR, SR, SPL, RGS, and RGSPL by 4.09\%, 4.41\%, 3.98\%, 3.07\%, and 2.28\%, respectively.
\subsection{Ablation Studies}
\begin{figure}[t]
    \centering
    \includegraphics[width=\linewidth]{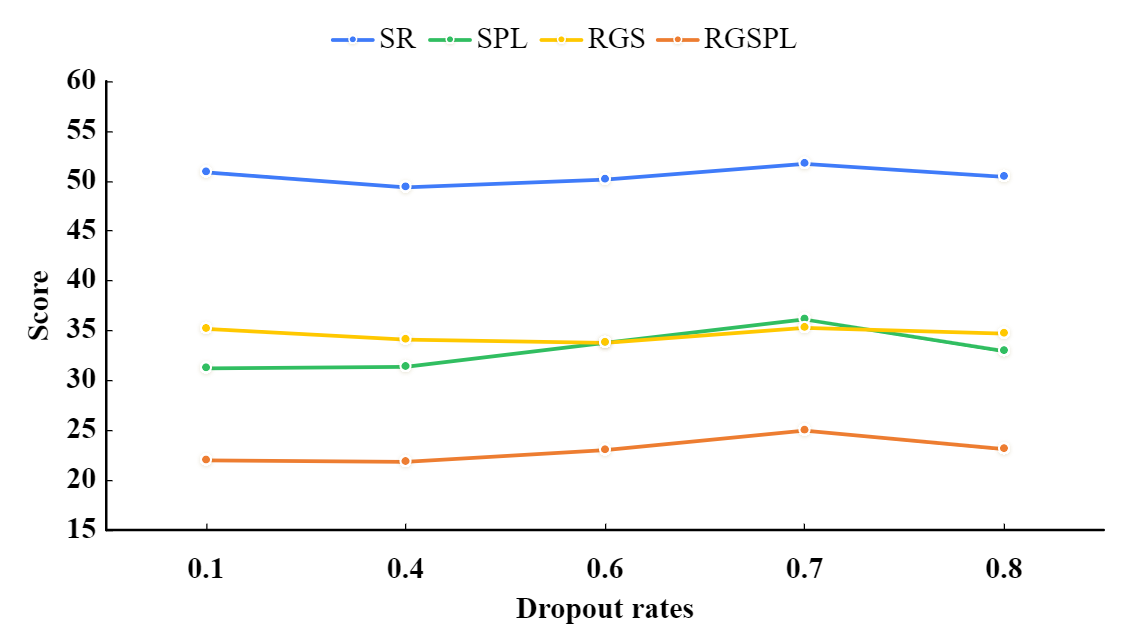} 
    \caption{Explanation of the impact of different Dropout rates on the metrics.}
    \label{fig:zhexiantu}
\end{figure}
\begin{figure*}[t]
    \centering
    \includegraphics[width=\textwidth]{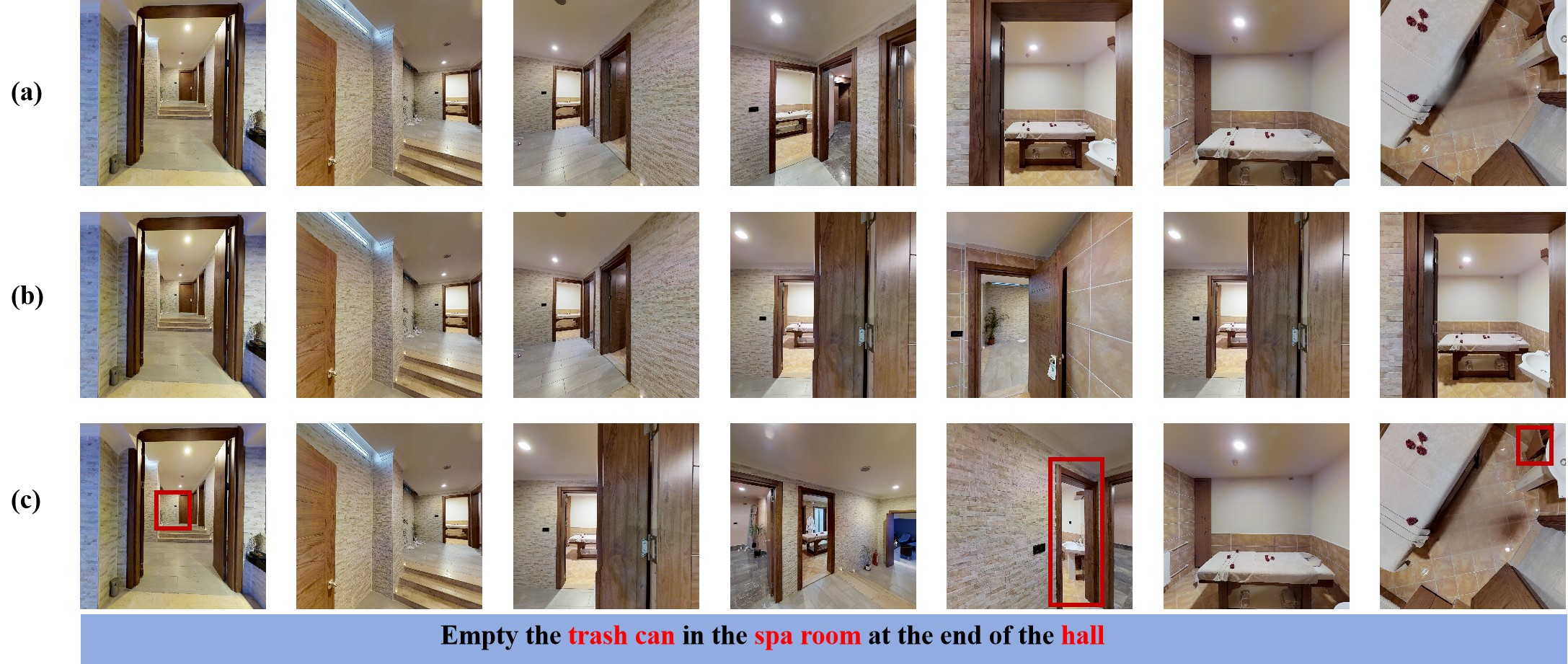} 
    \caption{Visualization of navigation examples. The red font and red boxes represent object information. (a) Ground truth navigation trajectory; (b) Navigation result of the baseline model (DUET), which only navigates to the SPA room but fails to locate the target object (trash can) correctly; (c) Navigation result of our model, successfully navigating to the SPA room and accurately locating the target object (trash can).}
    \label{fig:guijitu}
\end{figure*}
\label{subsec:ablation_studies}
We conducted ablation studies on the unseen validation set of the REVERIE dataset to evaluate the impact of each module on the model's performance. The results are shown in Table \ref{tab:all} and Table \ref{tab:past}.
\subsubsection{Ablation of Object Enhancement Modules}
\label{subsubsec:ablation_object_enhancement}
To assess the impact of the IOPA and TOPA modules on the agent's capabilities, we integrated them separately into the baseline model, with the results shown in Table \ref{tab:all}. When either the IOPA or TOPA module is used alone, all metrics outperform the baseline model. When both modules are used together, the improvements are more significant, especially in the SR and RGS metrics, which increase by 4.74\% and 3.12\%, respectively. This indicates that the object enhancement modules help the model better understand environmental information and instructions.
\subsubsection{Ablation of Object Representation Enhancement Modules}
\label{subsubsec:ablation_object_representation_enhancement}

To evaluate the impact of the OPE module on the IOPA and TOPA modules, we integrated the OPE module into the IOPA and TOPA modules separately, as shown in the results of Table \ref{tab:past}. Compared to the models without OPE, the IOPA module with OPE shows improvements in all metrics. For the TOPA module, all metrics except for the RGS indicator are improved after adding OPE. The complete model with OPE demonstrates significant improvements across all metrics compared to the IOPA and TOPA modules without OPE, indicating that OPE plays a crucial role in enhancing the performance of both the IOPA and TOPA modules.

\subsection{Quantitative Results}
\label{subsec:quantitativee_results}
Due to the relatively small scale of the REVERIE and R2R datasets, the model is prone to overfitting, which leads to a decrease in its generalization ability. To mitigate overfitting and enhance the model's generalization performance, we employed the Dropout method and experimentally investigated the impact of different Dropout hyperparameters on model performance. As shown in Figure \ref{fig:zhexiantu}, we tested five different Dropout rates (0.1, 0.4, 0.6, 0.7, 0.8) on the unseen subset of the REVERIE dataset. The experimental results show that the performance of all metrics initially decreases and then increases as the Dropout rate rises from 0.1 to 0.7. The best performance across all metrics is achieved at a Dropout rate of 0.7, after which the performance of all metrics starts to decrease. Therefore, we set the Dropout hyperparameter to 0.7.

\subsection{Qualitative Results}
\label{subsec:qualitative_results}
An example of navigation on the unseen validation set of the REVERIE dataset is illustrated in Figure \ref{fig:guijitu}, with the navigation process from left to right. The blue-boxed sentence is the instruction for this navigation, and the red words are the target objects we extracted. Figure \ref{fig:guijitu}(a) shows the ground truth navigation trajectory, requiring seven movements. Figure \ref{fig:guijitu}(b) displays the navigation result of the baseline model(DUET)\cite{chen2022think}, where the agent only navigates to the SPA room but fails to locate the target object correctly. Figure \ref{fig:guijitu}(c) shows the navigation result of our model. Compared to the baseline, our model successfully navigates to the target location (SPA room) and accurately locates the target object (trash can). This improvement is attributed to our Dual Object Perception-Enhancement, which strengthens the understanding of objects from both language and visual perspectives.
\section{Conclusion}
The Dual Object Perception-Enhancement Network (DOPE), proposed in this paper, effectively enhances language understanding and visual perception capabilities in Vision-and-Language Navigation (VLN) tasks. It achieves this by comprehensively integrating the Text Semantic Extraction (TSE), the Text Object Perception-Augmentation (TOPA), and the Image Object Perception-Augmentation (IOPA). Experimental results from the R2R and REVERIE datasets show that DOPE achieves superior navigation performance compared to existing methods, thus validating the effectiveness of our approach.

\section*{Acknowledgements}

This research was financially supported by the National Natural Science Foundation of China (Grant No. 62463029) and the Natural Science Foundation of Xinjiang Uygur Autonomous Region (Grant No. 2015211C288).

\bibliographystyle{ACM-Reference-Format}
\bibliography{sample-base}
\appendix
\end{document}